\documentclass[10pt,twocolumn,letterpaper]{article}

\usepackage{wacv}
\usepackage{times}
\usepackage{helvet}
\usepackage{courier}
\usepackage{xcolor}
\usepackage{graphicx}
\usepackage{amssymb}
\usepackage{amsmath}
\usepackage{algorithm}
\usepackage{algpseudocode}
\usepackage{url}
\usepackage{fancyhdr}
\usepackage{fancyvrb}
\DefineVerbatimEnvironment{code}{Verbatim}{fontsize=\small}
\DefineVerbatimEnvironment{example}{Verbatim}{fontsize=\small}

\wacvfinalcopy 

\def\wacvPaperID{486} 

\ifwacvfinal\fi
\fancypagestyle{firststyle}
{
   
   \fancyhf{}
   \fancyfoot[C]{\raggedright \footnotesize
         \copyright  2016 IEEE. Personal use of this material is permitted. Permission from IEEE must be obtained for all other uses, in any current or future media, including reprinting/republishing this material for advertising or promotional purposes, creating new
collective works, for resale or redistribution to servers or lists, or reuse of any copyrighted component of this work in other works. DOI: 10.1109/WACV.2016.7477571}
}

\begin{document}

\title{A Structured Approach to Predicting Image Enhancement Parameters}

\author{Parag Shridhar Chandakkar \hspace{2cm} Baoxin Li \\
School of Computing, Informatics and Decision Systems Engineering, Arizona State University\\
{\tt\small \{pchandak,baoxin.li\}@asu.edu}
}

\maketitle
\ifwacvfinal\thispagestyle{empty}\fi
\thispagestyle{firststyle}

\begin{abstract}
Social networking on mobile devices has become a commonplace of everyday life. In addition, photo capturing process has become trivial due to the advances in mobile imaging. Hence people capture a lot of photos everyday and they want them to be visually-attractive. This has given rise to automated, one-touch enhancement tools. However, the inability of those tools to provide personalized and content-adaptive enhancement has paved way for machine-learned methods to do the same. The existing typical machine-learned methods heuristically (e.g. $k$NN-search) predict the enhancement parameters for a new image by relating the image to a set of similar training images. These heuristic methods need constant interaction with the training images  which makes the parameter prediction sub-optimal and computationally expensive at test time which is undesired.  This paper presents a novel approach to predicting the enhancement parameters given a new image using only its features, without using any training images. We propose to model the interaction between the image features and its corresponding enhancement parameters using the matrix factorization (MF) principles. We also propose a way to integrate the image features in the MF formulation. We show that our approach outperforms heuristic approaches as well as recent approaches in MF and structured prediction on synthetic as well as real-world data of image enhancement.
\end{abstract}

\noindent \section{Introduction}

The growth of social networking websites such as Facebook, Google+, Instagram etc. along with the ubiquitous mobile devices has enabled people to generate multimedia content at an exponentially increasing rate. Due to the easy-to-use photo-capturing process of mobile devices, people are sharing close to two billion photos per day on the social networking sites \footnote{\url{http://www.kpcb.com/internet-trends}}. People want their photos to be visually-attractive which has given rise to automated, one-touch enhancement tools. However, most of these tools are pre-defined image filters which lack the ability of doing content-adaptive or personalized enhancement. This has fueled the development of machine-learning based image enhancement algorithms. 

Many of the existing machine-learned image enhancement approaches first learn a model to predict a score quantifying the aesthetics of an image. Then given a new low-quality image\footnote{We call the images before enhancement as low-quality and those after enhancement as high-quality in the rest of this article. The process of enhancing a new image is called ``the testing stage''.}, 
a widely-followed strategy to generate its enhanced \textit{version} is as follows:

\begin{itemize}
\item Generate a large number of candidate enhancement parameters\footnote{The brightness, saturation and contrast are referred to as ``parameters'' of an image in this article.} by densely sampling the entire range of image parameters. Computational complexity may be reduced by applying heuristic criteria such as, densely sampling only near the parameter space of most similar training images.
\item Apply these candidate parameters to the original low-quality image to create a set of candidate images.
\item Perform feature extraction on every candidate image and then compute its aesthetic score by using the learned model. 
\item Present the highest-scoring image to the user.
\end{itemize}

There are two obvious drawbacks for the above strategy. First, generating and applying a large number of candidate parameters to create candidate images may be computationally prohibitive even for low-dimensional parameters. For example, a space of three parameters where each parameter $\in \{0, . . . , 9\}$ produces $10^3$ combinations. Second, even if creating candidate images is efficient, extracting features from them is always computationally intensive and is the bottleneck. Also, such heuristic methods need constant interaction with the training database (which might be stored on a server) that makes the parameter prediction sub-optimal. All these factors contribute to making the testing stage inefficient.

Our approach assumes that a model quantifying image aesthetics has already been learned and instead focuses on finding a structured approach to enhancement parameter prediction. During training, we learn the inter-relationship between the low-quality images, its features, its parameters and the high-quality enhancement parameters. During the testing stage, we only have access to a new low-quality image, its features, parameters and the learned model and we have to predict the enhancement parameters. Using these enhancement parameters, we can generate the candidate images and select the best one using the learned model. The stringent requirement of not accessing the training images arises from real-world requirements. For example, to enhance a single image, it would be inefficient to establish a connection with the training database, generate hundreds of candidate images, perform feature extraction on them and then find the best image.

The search space spanned by the parameters is huge. However, the enhancement parameters are not randomly scattered. Instead they depend on the parameters and features of the original low-quality image. Thus we hypothesize that the enhancement parameters should have a low-dimensional structure in another latent space. We employ an MF-based approach because it allows us express the enhancement parameters in terms of three latent variables, which model the interaction across: 1. the low-quality images 2. their corresponding enhancement parameters 3. the low-quality parameters. The latent factors are learned during inference by Gibbs sampling. Additionally, we need to incorporate the low-quality image features since the enhancement parameters also depend on the color composition of the image, which can be characterized by the features. The feature incorporation in this framework is achieved by representing the latent variable which models the interaction across these images as a linear combination of their features, by solving a convex $\ell_{2,1}$-norm problem. We review the related work on MF as well as image enhancement in the following section.

\section{Related Work}

Development of machine-learned image enhancement systems has recently been an active research area of immense practical significance. Various approaches have been put forward for this task. We review those works which improve the visual appearance of an image using automated techniques. To encourage research in this field, a database named MIT-Adobe FiveK containing corresponding low and high-quality images was proposed in \cite{bychkovsky2011learning}. The authors also proposed an algorithm to solve the problem of global tonal adjustment. The tone adjustment problem only manipulates the luminance channel, where we manipulate saturation, brightness and contrast of an image.

Content-based enhancement approaches have been developed in the past which try to improve a particular image region \cite{berthouzoz2011framework,kaufman2012content}. These approaches require segmented regions which are to be enhanced. This itself may prove to be difficult. Approaches which work on pixels have also been developed using local scene descriptors. Firstly, similar images from the training set are retrieved. Then for each pixel in the input, similar pixels were retrieved from the training set, which were then used to improve the input pixel. Finally, Gaussian random fields maintain the spatial smoothness in the enhanced image. This approach does not consider the global information provided by the image and hence the enhancements may not be visually-appealing when viewed globally. In \cite{kang2010personalization}, a small number of image enhancements were collected from the users which were then used along with the additional training data.

Two recent works involving training a ranking model from low and high-quality images are presented in \cite{relativeLearningParag,yan2014learning}. The authors of \cite{yan2014learning} create a data-set of $1300$ corresponding low and high-quality image pairs along with a record of the intermediate enhancement steps. A ranking model trained on this type of data can quantify the aesthetics of an image. In \cite{relativeLearningParag}, non-corresponding low and high-quality image pairs extracted from the Web are used to train a ranking model. Both of these approaches use $k$NN-search during the testing stage to create candidate images. After extracting features and ranking them, the best image is presented to the user.

The task of enhancement parameter prediction could be related to the attribute prediction \cite{relativeAttributes,parikh2012relative,li2013relative,chen2014predicting}. However, the goal of the work on attribute prediction has been to predict relative strength of an attribute in the data sample (or image). We are not aware of any work which predicts parameters of an enhanced version of a low-quality image given only the parameters and features of that image. Since our approach is based on MF principles, we review the related recent work on MF.

MF \cite{rennie2005fast,mnih2007probabilistic,salakhutdinov2008bayesian,lawrence2009non,xiong2010temporal} is extensively used in recommender systems \cite{ma2008sorec,baltrunas2011matrix,ma2011recommender,Tang2015Exploring,marlin2012collaborative,song2015incremental,shi2014collaborative}. These systems predict the rating of an item for a user given his/her existing ratings for other items. For example, in Netflix problem, the task is to predict favorite movies based on user's existing ratings. MF-based solutions exploit following two key properties of such user-item rating matrix data. First, the preferred items by a user have some similarity to the other items preferred by that user (or by other similar users, if we have sufficient knowledge to build a similarity list of users). Second, though this matrix is very high-dimensional, the patterns in that that matrix are structured and hence they must lie on a low-dimensional manifold. For example, there are $17,770$ movies in Netflix data and ratings range from $1-5$. Thus, there are $5^{17770}$ rating combinations possible per user and there are $480,189$ users. Therefore, the number of actual variations in the rating matrix should be a lot smaller than the number of all possible rating combinations. These variations could be modeled by latent variables lying near a low-dimensional manifold. This principle is formalized in \cite{mnih2007probabilistic} with probabilistic matrix factorization (PMF). It hypothesizes that the rating matrix can be decomposed into two latent matrices corresponding to user and movies. Their dot product should give the user-ratings. This works fairly well on a large-scale data-set such as Netflix. However, a lot of parameters have to be tuned. This requirement is alleviated in \cite{salakhutdinov2008bayesian} by developing a Bayesian approach to MF (BPMF). BPMF has been extended for temporal data (BPTF) in \cite{xiong2010temporal}. MF is used in other domains such as computer vision to predict feature vectors of another viewpoint of a person given a feature for one viewpoint \cite{chen2014inferring}. We adopt and modify BPTF since it allows us to model joint interaction across low-quality images, corresponding enhancement parameters and the low-quality parameters. In the next section, we detail our problem formulation and proposed approach.


\section{Problem Formulation and Proposed Approach}
We have a training set consisting of $N$ images $\{\textbf{S}_1,\hdots,\textbf{S}_N\}$\footnote{We use bold letters to denote matrices. Non-bold letters denote scalars/vectors which will either be clear from the context or will be mentioned. $X^i , X_i , \mathbf{X}^T , X_{ij}$ and $||\mathbf{X}||_p$ denote row, column, transpose, entry at row $i$ and column $j$ of a matrix $\mathbf{X}$ and $p^{th}$ norm of matrix $\mathbf{X}$ respectively.}. Parameters of all images are represented as $\mathbf{A}=\{A_1,\hdots,A_N\}$ where $A_i \in \mathbb{R}^{K \times 1} \ \forall \ i \in \{1,\hdots,N\}$. Each image has $M$ enhanced versions and each version has the same size as that of its corresponding low-quality image. All versions corresponding to the $i^{th}$ image are represented as $\{\textbf{W}_i^1,\hdots,\textbf{W}_i^M\}$. All versions are of higher quality as compared to its corresponding image. Parameters of all $M$ versions of the $i^{th}$ image (also called as candidate parameters) are represented as $\mathbf{A}^\prime=\{{A^\prime_i}^1,\hdots,{A^\prime_i}^M\}$, where ${A^\prime}_i^j \in \mathbb{R}^{K \times 1} \ \forall \ i,j$. Features of all low-quality images are represented as $\mathbf{F}=\{F_1,\hdots,F_N\}$ where $F_i \in \mathbb{R}^{L \times 1} \ \forall \ i$. In practice, we observe that $M \ll N, K < M$. Our goal is to be able to predict the candidate parameters for all the versions of the $i^{th}$ image by only using the information provided by $A_i$ and $F_i$. To the best of our knowledge, this is a novel problem of real significance that has not been addressed in the literature. We now explain our proposed approach.

As mentioned before, our task is to predict the candidate parameters for all the enhanced versions of a low-quality image with the help of its parameters and features. The values for all the $K$ parameters corresponding to $N$ images and their $N \cdot M$ versions (total $N+N \cdot M)$ can be stored in three-dimensional matrix $\mathbf{R} \in \mathbb{R}^{N \times (M+1) \times K}$. We need to predict $\hat{R}^k_{ij}=R_i^k+\Delta R_{ij}^k$ or in turn just $\Delta R_{ij}^k$. $R_i^k$ denotes the $k^{th}$ parameter value ($k \in \{1,\hdots,K\})$ of the $i^{th}$ low-quality image and $\hat{R}_{ij}^k$ is the $k^{th}$ parameter value of $j^{th}$ version of the $i^{th}$ image. Given a new $n^{th}$ low-quality image, we only need to predict $\Delta R_{nj}^k \ \forall \ j=\{1,\hdots,M\}, \forall \ k$. 

During training, we can compute $\Delta R_{ij}^k$ from available $R_{ij}^k$ and $\hat{R}^k_{ij}$. Following MF principles, we express $\mathbf{\Delta R}$ as an inner product of three latent factors, $\mathbf{U} \in \mathbb{R}^{D \times N}, \mathbf{V} \in \mathbb{R}^{D \times M}$ and $\mathbf{T} \in \mathbb{R}^{D \times K}$ \cite{salakhutdinov2008bayesian,xiong2010temporal}. $D$ is the latent factor dimension. These latent factors should presumably model the underlying low-dimensional subspace corresponding to the low-quality images, its enhanced versions and its parameters. This can be formulated as:

\begin{equation}\label{eq:parafacModel}
\Delta R_{ij}^k=<U_i,V_j,T_k>  \equiv \sum \limits_{d=1}^D U_{di} V_{dj} T_{dk},
\end{equation}

\noindent where $U_{di}$ denotes the $d^{th}$ feature of the $i^{th}$ column of $\mathbf{U}$. Presumably, as we increase $D$, the approximation error $\Delta R_{ij}^k-<U_i,V_j,T_k>$ should decrease (or stay constant) if the prior parameters for latent factors $\mathbf{U},\mathbf{V}$ and $\mathbf{T}$ are chosen correctly. Following \cite{salakhutdinov2008bayesian}, we choose normal distribution (with precision $\alpha$) for: 1. the conditional distribution $\mathbf{\Delta R} | (\mathbf{U},\mathbf{V},\mathbf{T})$ and 2. for prior distributions - $p(\mathbf{U} | \Theta_U), p(\mathbf{V} | \Theta_V)$ and $p(\mathbf{T} | \Theta_T)$, where $\Theta_U=(\mathbf{\mu}_U,\mathbf{\Lambda}_U^{-1})$, $\Theta_V=(\mathbf{\mu}_V,\mathbf{\Lambda}_V^{-1})$, $\Theta_T=(\mathbf{\mu}_T,\mathbf{\Lambda}_T^{-1})$. $\Theta_U, \Theta_V$ and $\Theta_T$ are hyper-parameters, and $\mu$ and $\mathbf{\Lambda}$ are the multivariate precision matrix and the mean respectively. Since the Wishart distribution is a conjugate prior for multivariate normal distribution (with precision matrix), we put Gaussian-Wishart priors on all hyper-parameters\footnote{For details, see supplementary material on author's website.}. We could find the latent factors $\mathbf{U},\mathbf{V}$ and $\mathbf{T}$ by doing inference through Gibbs sampling. It will sample each latent variable from its distribution, conditional on the values of other variables. The predictive distribution for $\Delta R_{ij}^k$ can be found by using Monte-Carlo approximation (explained later). 

However, it is important to note the following major differences in our problem when compared with the previous work on MF \cite{salakhutdinov2008bayesian,xiong2010temporal}. In product or movie rating prediction problems, an average (non-personalized) recommendation may be provided to a user who has not provided any preferences (not necessarily constant for all users). In our case, each image may require a different kind of parameter adjustment to create its enhanced version and thus no ``average'' adjustment exists. As explained before, the adjustment should depend on the image's features, which characterize that image (e.g. bright vs. dull, muted vs. vibrant). In our problem, it is particularly difficult to get a good generalizing performance on the testing set as we shall see later. The loss in performance of existing approaches on the testing set can be attributed to the different requirements for parameter adjustments for each image. Thus it becomes necessary to include the information obtained from image features into the formulation. We show that simply concatenating the parameters and features and applying MF techniques presented in \cite{salakhutdinov2008bayesian,xiong2010temporal} does not provide good performance, possibly because they lie in different regions of the feature space.

%

To overcome this problem, we observe that the conditional distribution of each $U_i$ factorizes with respect to the individual samples. We propose to express $\mathbf{U}$ as a linear function of $\mathbf{F}$ by using a convex optimization scheme. We then integrate it into the inference algorithm to find out the latent factors. The linear transformation can be expressed as,

\begin{equation} \label{eq:U_decomposition}
U_i=F_i^T \mathbf{P}+Q, \ \forall \ i \in \{1,\hdots,N\},
\end{equation}

\noindent where $F_i \in \mathbb{R}^{L \times 1}, U_i \in \mathbb{R}^{D \times 1}, \mathbf{P} \in \mathbb{R}^{D \times D}$ and $Q \in \mathbb{R}^{1 \times D}$. Note that to carry out this decomposition, we have to set $D=L$. This is not a severe limitation since $L$ is usually large ($\sim 1000$) and as we have mentioned before, increasing $D$ should decrease the approximation error at the cost of increased computation. Henceforth we assume that our feature extraction process generates $F_i \in \mathbb{R}^{D \times 1}$. Also, note that large $L$ does not mean that the latent space is no longer low-dimensional, because $L$ is still smaller as compared to all the possible combinations of parameters (e.g. $5^{17770}$).

We propose an iterative convex optimization process to determine coefficients $\mathbf{P}$ and $Q$ of Equation \ref{eq:U_decomposition}. We propose the following objective function to determine them:
q
\begin{multline} \label{eq:objectiveFunction}
\min_{\mathbf{P},Q} \sum \limits_{i=1}^N ||F_i^T \mathbf{P} + Q - U_i^T ||_2 + \beta ||\mathbf{P}||_{2,1} + \gamma ||Q||_2
\end{multline}

The objective function tries to reconstruct $\mathbf{U}$ using $\mathbf{P},Q$ and $F$ while controlling the complexity of coefficients. Let's concentrate on the structure of $\mathbf{P}$ (by neglecting the effect of $Q$ momentarily). The columns of $\mathbf{P}$ act as coefficients for $F_i$. Ideally, we would want the elements of $U_i$ to be determined by a sparse set of features, which implies sparsity in the columns of $\mathbf{P}$. To this end, we impose $\ell_{2,1}$-norm on $\mathbf{P}$, which gives us a block-row structure for $\mathbf{P}$.

Let us consider the structure of $Q$ along with $\mathbf{P}$. Equation \ref{eq:U_decomposition} shows that different columns of $U_i$ depend on different image features $F_i$. Also, we expect that a different set of columns of $\mathbf{P}$ will get activated (take on large values) for different $F_i$. We add an offset $Q \in \mathbb{R}^{1 \times D}$ for regularization. Thus the offset introduced by $Q$ remains constant across all the images but changes for each $F_{i,j}$. Making $Q$ to be a row vector also forces $\mathbf{P}$ to play a major role in Equation \ref{eq:objectiveFunction}. This in turn increases the dependence of $U_i$ on $F_i$. If we were to define $Q$ as the same size of $\mathbf{U}$ (which would mean different offsets for each image as well as its features), it would pose two potential disadvantages. Firstly, optimal $\mathbf{P}$ and $Q$ could be (trivially) obtained by just setting each entry of $\mathbf{P}$ to a very small value and letting a column of $Q \approx U_i$ (which makes $F_i$ redundant). Secondly, while testing for a new image, we would have to devise a strategy to determine the suitable value for $Q$. For example, we could take the column of $Q$ that corresponds to the nearest training image. This adds unnecessary complexity and reduces generalization. By making $Q$ a row vector, we consider that it may be possible to arrive to the space of enhancement parameters by linearly transforming the low-quality image features with a constant offset. In other words, we want $\mathbf{P}$ to transform the features into a region in the latent space where all the other high-quality images lie and $Q$ provides an offset to avoid over-fitting. This is a joint $\ell_{2,1}$-norm problem which can be solved efficiently by reformulating it as convex. We thus reformulate Equation \ref{eq:objectiveFunction} as follows, inspired by \cite{feipingNe}:

\begin{equation} \label{eq:reformulation1}
\min_{\mathbf{P},Q} \ \frac{1}{\beta} \sum \limits_{i=1}^N ||F_i^T \mathbf{P} + Q - U_i^T ||_2 + ||\mathbf{P}||_{2,1} + \frac{\gamma}{\beta} ||Q||_2
\end{equation}

The $\ell_{2,1}$-Norm of a matrix $\mathbf{X} \in \mathbb{R}^{M \times N}$ is defined as, $\ell_{2,1}(\mathbf{X}) = \sum \limits_{i=1}^M ||\mathbf{X}^i||_2$. Also, for a row vector $Q$, we have $||Q||_2=||Q||_{2,1}$. Thus Equation \ref{eq:reformulation1} can be further written as:

\begin{equation} \label{eq:reformulation2}
\min_{\mathbf{P},Q} \ \frac{1}{\beta} ||\mathbf{F}^T \mathbf{P}+1^N Q-\mathbf{U}^T||_{2,1}+||\mathbf{P}||_{2,1}+\delta ||Q||_{2,1},
\end{equation}

\noindent where $\delta=\frac{\gamma}{\beta}$ and $1^N$ is a column vector of ones $\in \mathbb{R}^N$. Now, put $\mathbf{F}^T \mathbf{P} + 1^N Q - \beta \mathbf{E} = \mathbf{U}^T$. Thus Equation \ref{eq:reformulation2} becomes:

\begin{align} 
\begin{split} \label{eq:reformulation3}
&\hspace{-190pt}\min_{\mathbf{E},\mathbf{P},Q} \ ||\mathbf{E}||_{2,1} + ||\mathbf{P}||_{2,1} + \delta ||Q||_{2,1},\\
&\hspace{-187pt}\text{s.t.} \ \ \mathbf{F}^T \mathbf{P} + 1^N Q - \beta \mathbf{E} = \mathbf{U}^T, \\
\hspace{-38pt}\min_{\mathbf{E},\mathbf{P},Q} \ \left\vert\left\vert \begin{bmatrix}
        \mathbf{E} \\
        \mathbf{P} \\
        \delta Q 
    \end{bmatrix} \right\vert\right\vert_{2,1} \text{s.t.} \ \left[-\beta \mathbf{I}_N \ \ \mathbf{F}^T \ \ \delta^{-1} 1^N \right] \begin{bmatrix}
        \mathbf{E} \\
        \mathbf{P} \\
        \delta Q 
    \end{bmatrix} = \mathbf{U}^T 
\end{split}
\end{align}


\begin{algorithm}[!t]
\caption{\hspace{4.5pt} Gibbs Sampling for Latent Factor Estimation}
\label{algo:CHalgorithm}
\begin{algorithmic}
\Statex Initialize model parameters $\{\mathbf{P}^{(1)},Q^{(1)},\mathbf{V}^{(1)},\mathbf{T}^{(1)}\}$. 
Obtain $\left(\mathbf{U}^{(1)}\right)^T=\mathbf{F}^T \mathbf{P}^{(1)}+Q^{(1)}$ \\
\vspace{6pt}
For $y=1,2,\hdots,Y$ \\

\begin{itemize}
\item Sample the hyper-parameters according to the derivations \footnotemark:
\end{itemize}
\hspace{15.5pt} $\alpha^{(y)} \sim p(\alpha^{(y)}| \mathbf{U}^{(y)},\mathbf{V}^{(y)},\mathbf{T}^{(y)},\Delta \mathbf{R})$, \\
\hspace{35pt} $\Theta_U^{(y)} \sim p(\Theta_U^{(y)} | \mathbf{U}^{(y)})$, \ \ $\Theta_V^{(y)} \sim p(\Theta_V^{(y)} | \mathbf{V}^{(y)})$, $\Theta_T^{(y)} \sim p(\Theta_T^{(y)} | \mathbf{T}^{(y)})$\\

\vspace{6pt}
\begin{itemize}
\item For $i=1,...,N$, sample the latent features of an image (in parallel):
\end{itemize}
\hspace{17.5pt} $U_i^{(y+1)} \sim  p(U_i|\mathbf{V}^{(y)},\mathbf{T}^{(y)},\Theta_U^{(y)},\alpha^{(y)},\Delta \mathbf{R})$
\vspace{2pt}
\\ \hspace{9pt} Determine $\mathbf{P}^{(y+1)}$ and $Q^{(y+1)}$ using the iterative \\\hspace{9.2pt} optimization by substituting $\mathbf{B}=\left(\mathbf{U}^{(y+1)}\right)^T$.
\\ Reconstruct $\mathbf{U}^{(y+1)}$: $\left(\hat{\mathbf{U}}^{(y+1)}\right)^T=\mathbf{F}^T \mathbf{P}^{(y+1)} + Q^{(y+1)}$

\begin{itemize}
\item For $j=1,...,M$, sample the latent features of the enhanced versions (in parallel):
\end{itemize}
\hspace{20pt} $V_j^{(y+1)} \sim  p(V_j|\hat{\mathbf{U}}^{(y+1)},\mathbf{T}^{(y)},\Theta_V^{(y)},\alpha^{(y)},\Delta \mathbf{R})$

\begin{itemize}
\item For $k=1,...,K$, sample the latent features of parameter (in parallel):
\end{itemize}
\hspace{20pt} $T_k^{(y+1)} \sim  p(T_k | \hat{\mathbf{U}}^{(y+1)},\mathbf{V}^{(y+1)},\Theta_T^{(y)},\alpha^{(y)},\Delta \mathbf{R})$

\end{algorithmic}
\end{algorithm}


\noindent Equation \ref{eq:reformulation3} is now in the form of: $\displaystyle \min_\mathbf{X} \ ||\mathbf{X}||_{2,1} \ \ \text{s.t.} \ \ \mathbf{ZX}=\mathbf{B}$ and thus convex. It can be iteratively solved by an efficient algorithm mentioned in \cite{feipingNe}. We set $\beta=0.1$ and $\delta=3$. Once we have expressed $\mathbf{U}$ as a function of $\mathbf{F}$, we use Gibbs Sampling to determine the latent factors $\mathbf{P},Q,\mathbf{V}$ and $\mathbf{T}$ \cite{salakhutdinov2008bayesian}. As mentioned before, the predictive distribution for a new parameter value $\Delta \hat{R}_{ij}^k$ is given by a multidimensional integral as: 

\begin{equation}
\begin{aligned}
p(\Delta \hat{R}_{ij}^k | \Delta \mathbf{R})=&\int p(\Delta \hat{R}_{ij}^k | U_i,V_j,T_k,\alpha) \cdot \\& p(\mathbf{U},\mathbf{V},\mathbf{T},\alpha,\Theta_U,\Theta_V,\Theta_T | \Delta \mathbf{R}) \cdot \\ & d(\mathbf{U},\mathbf{V},\mathbf{T},\alpha,\Theta_U,\Theta_V,\Theta_T).
\end{aligned}
\end{equation}

We resort to numerical approximation techniques to solve the above integral. To sample from the posterior, we use Markov Chain Monte Carlo (MCMC) sampling. We use the Gibbs sampling as our MCMC algorithm. We can approximate the integral by,

\footnotetext{See supplementary material on author's website for detailed derivations.}

\begin{equation}
p(\Delta \hat{R}_{ij}^k | \Delta \mathbf{R}) \approx \sum \limits_{y=1}^Y p \left(\Delta \hat{R}_{ij}^k | U_i^{(y)}, V_j^{(y)}, T_k^{(y)}, \alpha^{(y)}\right)
\end{equation}

%

Here we draw $Y$ samples and the value of $Y$ is set by observing the validation error. The sampling from $\mathbf{U,V}$ and $\mathbf{T}$ is simple since we use conjugate priors for the hyper-parameters. Also, a random variable can be sampled in parallel while fixing others which reduces the computational complexity. Algorithm $1$ shows how to iteratively sample $\mathbf{U,V,T}$ and obtain $\mathbf{P}$ and $Q$. Note that it is required in the algorithm to reconstruct $\mathbf{U}^{(y+1)}$ at every iteration since there will always be a small reconstruction error $||\hat{\mathbf{U}}^{(y+1)}-\mathbf{U}^{(y+1)}||$. The error occurs because we force $Q$ to be a row vector, which makes the exact recovery of $\mathbf{U}^{(y+1)}$ difficult. The reconstructed error causes adjustment of $\mathbf{V}$ and $\mathbf{T}$. Once we obtain the four latent factors, our task is to predict the parameter values for $M$ enhanced versions having $K$ parameters each. Suppose $F_t$ is the feature vector of a new image, then the parameter values $\Delta \hat{R}_{tj}^k$ can be simply obtained by computing, $\Delta \hat{R}_{tj}^k=<F_t^T \mathbf{P} + Q, V_j,T_k> \ \forall \ j \in \{1,\hdots,M\}$ and $k \in \{1,\hdots,K\}$. If the parameter value predictions lie beyond a certain range then a thresholding scheme can be used based on the prior knowledge. For example, to constrain the predictions between $[0,1]$, a logistic function may be used.

\section{Experiments}

We conduct two experiments to show the effectiveness of our approach. We did the first one on a synthetic data and compared it with: 1. BPMF 2. our own discrete version of BPTF, called D-BPTF. 3. multivariate linear regression (MLR) 4. twin Gaussian processes (TGP) \cite{twinGP} 5. Weighted $k$NN regression (WKNN). For D-BPTF, we make minor modifications in the original BPTF approach \cite{xiong2010temporal} by removing the temporal constraints on their temporal variable, since there are no temporal constraints in our case. The inference for their temporal variable is then done in the exactly same manner as the other non-temporal variables. This gave us a marginal boost in the performance. For MLR, We use a standard multivariate regression by maximum likelihood estimation method. Specifically, we use MATLAB's \url{mvregress} command. TGP is a generic structured prediction method. It accounts correlation between both input and output resulting in improved performance as compared to MLR or WKNN. The WKNN approach predicts the test sample as a weighted combination of the $k$-nearest inputs. The first two algorithms do not allow features inclusion. For MLR, TGP and WKNN, we concatenate $A_i$ and $F_i$, and use it to predict ${A_i^\prime}^j$. Even for our approach, we concatenate $A_i$ and sample feature to form $F_i$. The intuition behind this concatenation is that the enhancement parameters should be a function of input parameters as well along with the features. We did observe performance boost after concatenating the features and parameters.

\begin{figure*}[!t]
\centering
\includegraphics[width=0.99\textwidth]{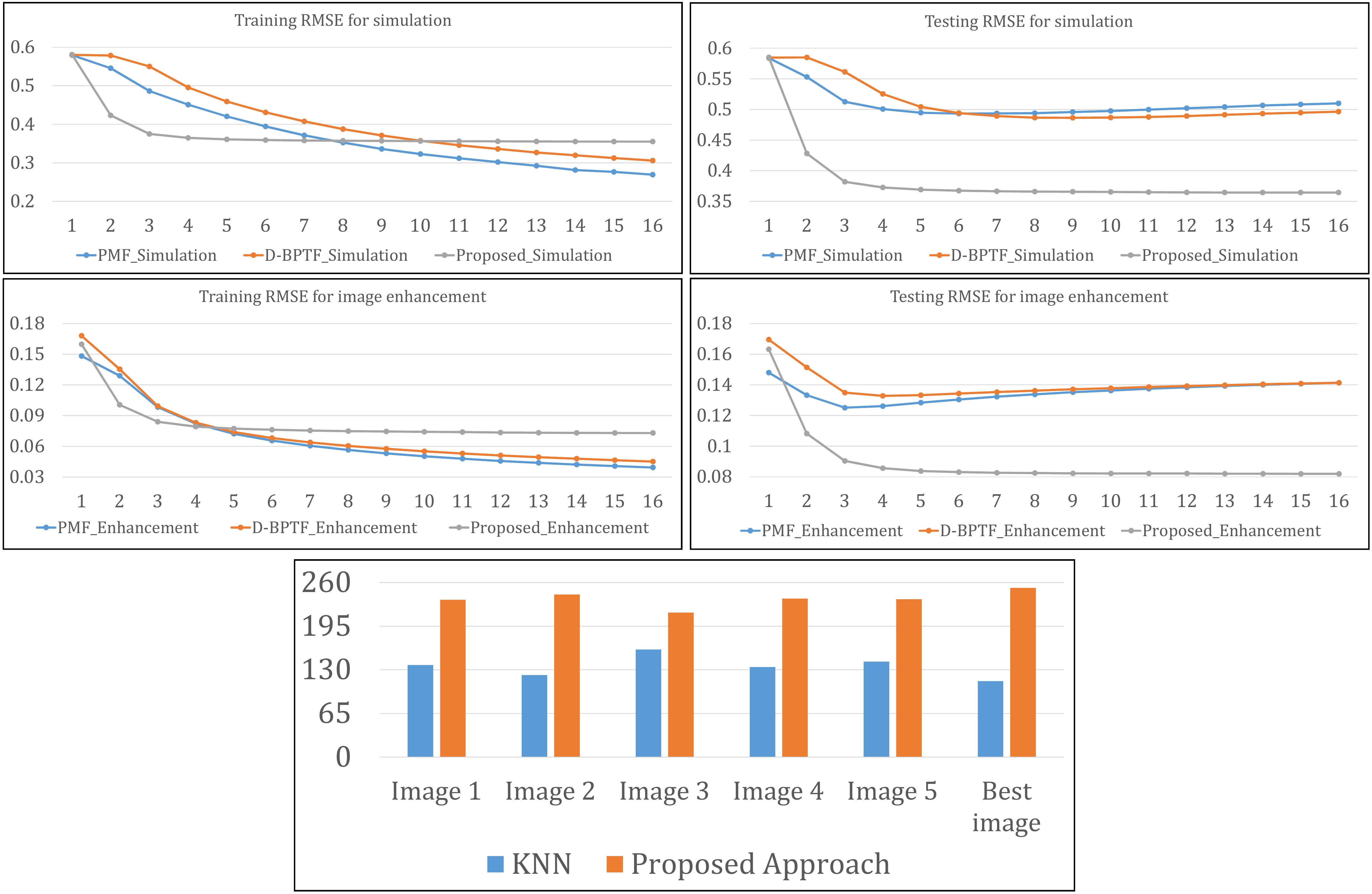}
\caption{Top plots: train and test RMSEs for both the experiments. Bottom plot: First 5 sets of bars show votes for version 1 to 5 of $k$NN vs. the best image of our approach. The last set of bars shows votes for the best image of both approaches. Please zoom in for better viewing. See in color.}
\label{fig:allPlots}
\end{figure*}

The second experiment demonstrates the usefulness of this approach in a real-world setting where we have to predict paramters of the enhanced versions of an image (then generate those versions by applying predicted parameters to the input low-quality image) without using any information about the versions. We compare our approach with the competing $5$ algorithms in addition to $k$NN-search as it is also used in \cite{learningToRank,relativeLearningParag}. We also analyzed the effect of $Q$ in our solution by: removing $Q$ i.e. $\mathbf{U}=\mathbf{F}^T \mathbf{P}$. 

\subsection{Data set description and experiment protocol}

\begin{figure*}[!t]
\centering
\includegraphics[width=0.99\textwidth]{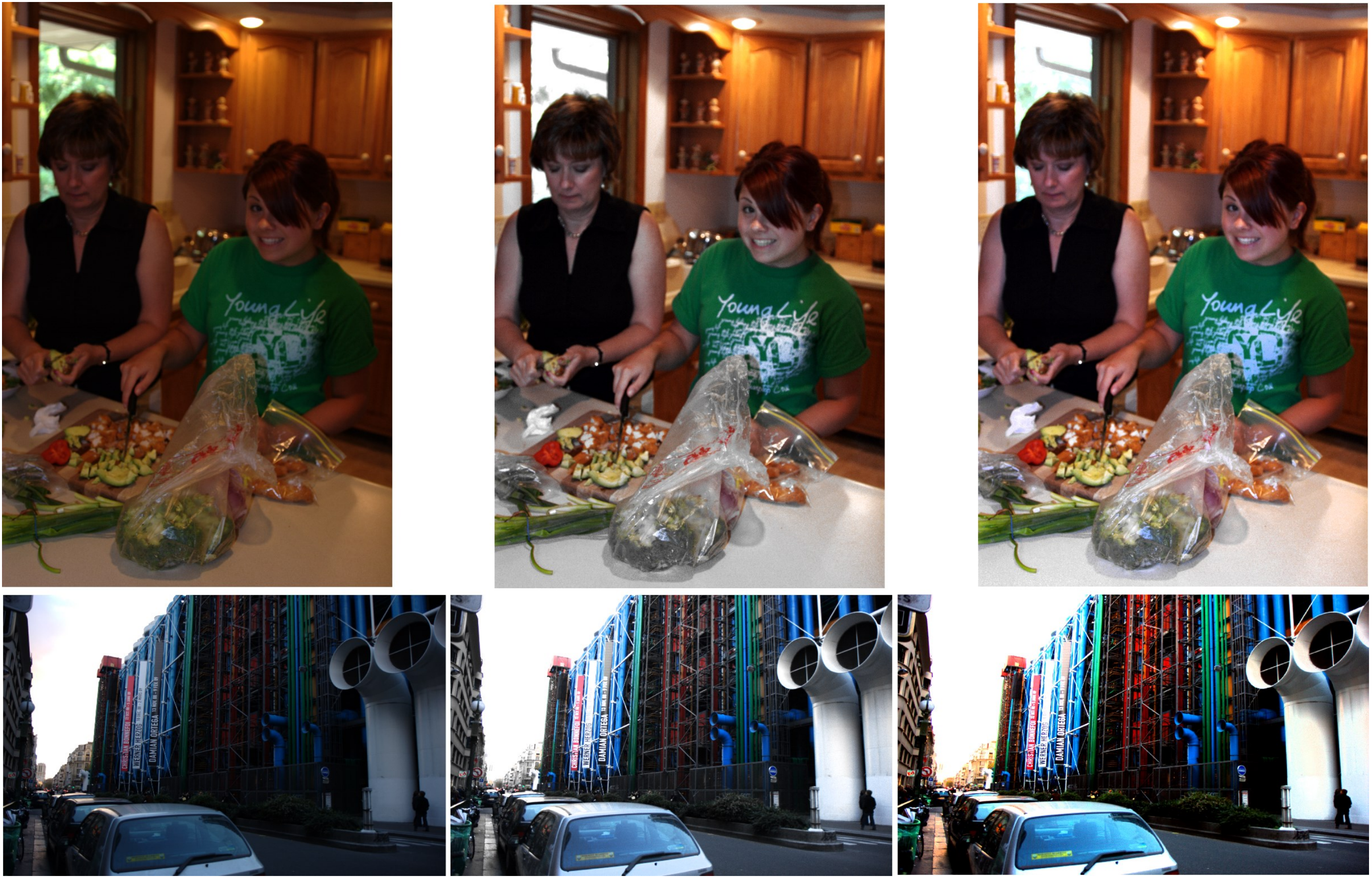}
\caption[]{Left: Original image, Middle: enhanced image by $k$NN and Right: proposed approach \footnotemark{}. See in color.}
\label{fig:enhancedImages}
\end{figure*}

The synthetic data is carefully constructed by keeping the following task in mind. We are given a training set consisting of: 1. $\mathbf{F} \in \mathbb{R}^{D \times N}$;  2. $\mathbf{A} \in \mathbb{R}^{K \times N}$; and 3. \textit{only} parameters of $M$ versions for each input sample - $\mathbf{A}^\prime \in \mathbb{R}^{K \times N \times M}$. Our aim is to predict parameters for a set of $M$ versions given a new $F_i$ and $A_i$. In real-world problems, $\mathbf{A}$ and $\mathbf{F}$ are interdependent. The parameters of $M$ versions are dependent on both $\mathbf{A},\mathbf{F}$. Hence we construct the synthetic data as follows. 

Firstly, we generate a set of $3$-D input parameters - $\mathbf{A}$ - drawn from a uniform distribution $[0,1]$. Then we generate a $50$-D feature set $\mathbf{F}$, where each element of $F_i$ is related to all $A_{k,i} \ \forall \ i=\{1,\hdots,10^3\}, k=\{1,2,3\}$ by a nonlinear function. For example, $F_{j,i}=r_1^{A_{1,i}} + \frac{1}{1+e^{-r_2 A_{2,i}}} + A_{3,i}^{r_3}, \forall \ j \in \{1,\hdots,50\}$ and $r_1,r_2,r_3$ are random numbers. The parameters of enhanced versions, $A_{k,i,m}^\prime$, are also non-linearly related to $A_{k,i} \ \forall \ k, \forall \ m \in \{1,\hdots,4\}$ and $F_i$. For example, $A_{k,i,m}^\prime=\eta\left(r_1^{A_{1,i}} + \frac{1}{1+e^{-r_2 A_{2,i}}} + A_{3,i}^{r_3}\right)+(1-\eta) \cdot ||F_i||_2$. The contribution of $F_i$ is decided by $\eta$. We perform 3-fold cross-validation. We predict the values of $\mathbf{A}^\prime$ in the test set (disjoint from training) using corresponding $\mathbf{A}$ and $\mathbf{F}$. RMSE is computed between the predicted and actual $\mathbf{A}^\prime$.

The MIT-Adobe FiveK data-set contains 5000 high-quality photographs taken with SLR cameras. Each photo is then enhanced by five experts to produce 5 enhanced versions. We extract average saturation, brightness and contrast for every image, which are parameters $\in \mathbf{A}$. We also extract $1274$-D color histogram with 26 bins for hue, 7 bins each for saturation and value. We also calculate localized features of $144$-D each for contrast, brightness and saturation. Finally, we append average saturation, brightness and contrast of the input low-quality image, which are its parameters. Thus we get a $1709$-D $(=1274+3 \times 144+3)$ representation for every image $\in \mathbf{F}$. We train using 4000 images and use 500 images each for validation and testing. We predict parameters for 5 versions in a $3 \times 5$ matrix for each image in the testing set. An entry $A_{i,j}^\prime$ denotes the value for $i^{th}$ parameter of $j^{th}$ enhanced version. To enable comparison with the expert-enhanced images of the data-set, we also compute parameters for 5 enhanced versions for each image, which we treat as ground-truth. We evaluate this experiment in two ways. Firstly, we calculate RMSE between the parameters of 5 expert-enhanced photos and the parameters of the predicted versions using five aforementioned algorithms. Secondly, we conduct a subjective test under standard test settings (constant lighting, position, distance from the screen). In this case, we compare our approach with the popular $k$NN-search-based approach. It first finds the nearest original image in the training set to the testing image - $\mathbf{im}$ - and then applies the same parameter transformation to $\mathbf{im}$ to generate $5$ version. In our approach, we predict the parameters for enhanced versions using the proposed formulation. We threshold the parameter values as: 

\begin{equation} \label{eq: clippingScheme}
\begin{aligned}
A_{k,i,m}^\prime&=\min(A_{k,i,m}^\prime,A_{k,i}+\lambda_k A_{k,i}), \\ A_{k,i,m}^\prime&=\max(A_{k,i,m}^\prime,A_{k,i}-\zeta_k A_{k,i}),
\end{aligned} 
\end{equation}

\noindent where $\mathbf{\lambda}$ and $\mathbf{\zeta}$ are multipliers for the $k^{th}$ parameter. In our case, the multipliers for saturation, brightness and contrast are: $\mathbf{\lambda}=\{0.4,0.4,0.05\}, \mathbf{\zeta}=\{0.3,0.3,0.01\}$. As mentioned before, the clipping scheme in our formulation should be set using prior knowledge. Here, we know that the enhanced images usually have a larger increase (as compared to decrease) associated with their parameters. Also, changing contrast by a very small amount affects the image greatly.

\footnotetext{See supplementary on author's website for additional full-resolution results.}

The predicted parameters are applied to the input image to obtain its enhanced versions. The procedure is the same for both the approaches and is as follows. First we change contrast till the difference between the updated and the predicted contrast is marginal. We update contrast first since changing it updates both brightness and saturation. We then update brightness and saturation till they come significantly closer to their corresponding predicted values. This gives us 5 versions for both approaches. To allow comparisons within a reasonable amount of time, we use a pre-trained ranking weight vector $w$ (from \cite{relativeLearningParag}) to select the best image of our approach (\textit{im-proposed}) and $k$NN-approach (\textit{im-$k$NN}). For the subjective test, people are told to compare \textit{im-proposed} with the 5 enhanced versions of $k$NN-approach as well as with \textit{im-$k$NN}. Thus for every input image, people perform 6 comparisons. The image order was randomized. We conducted the test with 11 people and 35 input images. Thus every person compared $210$ pairs of images. They were told to choose a visually-appealing image. The third option of simultaneously preferring both images was also provided. This option has no effect on cumulative votes.

\subsection{Results}

The parameters for the synthetic data were more accurately predicted by our approach than BPMF, D-BPTF, MLR, TGP and WKNN. It is worth noting that though the training error continues to decrease for our approach, BPMF and D-BPTF, the testing error starts increasing after only 5 and 8 iterations for BPMF and D-BPTF, respectively. However, testing error in our approach decreases rapidly for 4 iterations and then it decreases very slowly for the next 12, as shown in Fig. \ref{fig:allPlots}. The RMSE on test set for BPMF, D-BPTF, MLR, TGP, WKNN and the proposed approach is $0.4933, 0.4865, 0.6293, 0.4947, 0.8014$ and $0.3644$. The numbers show that our approach is able to effectively use the additional information provided by features and the interaction between $\mathbf{A},\mathbf{F}$ and all versions to provide better prediction. On the other hand, BPMF and D-BPTF start over-fitting quickly due to lack of sample feature information while MLR and WKNN fail to model the complex interaction between variables. TGP performs better because of its ability to capture correlations between input and output. However, TGP still treats each version independently and thus its performance still falls short of our approach.

In the second experiment, the RMSE for BPMF, D-BPTF, MLR, TGP, WKNN and our approach is $0.1251, 0.1328, 1.2420, 0.1268, 0.1518$ and $0.0820$ respectively. The testing error starts increasing after only 3 and 5 iterations for BPMF and D-BPTF, respectively. It is important to note that we do \textit{not} use the clipping scheme mentioned in Equation \ref{eq: clippingScheme} in order to do a fair comparison of RMSEs between all the five approaches and the proposed appraoch. For the subjective evaluation, Fig. \ref{fig:allPlots} shows cumulative votes obtained for ours and the $k$NN-based approach for comparison between 5 images chosen by $k$NN and the best image chosen by our approach. Fig. \ref{fig:allPlots} also shows votes obtained for the best images chosen by both approaches. Fig. \ref{fig:enhancedImages} shows two input images enhanced by both the approaches. The top row of Fig. \ref{fig:enhancedImages} shows that $k$NN reduces the saturation while increasing the brightness. Our approach balances both of them to obtain a more appealing image. In the bottom row, however, both approaches fail to produce aesthetic images as images become too bright. It is probably due to the portion of the sky in the input image. For both the images, most people prefer images enhanced by our approach. Computationally, our approach is superior than $k$NN. Complexity of our approach is independent of data-set size at testing time whereas $k$NN searches the entire data-set for the closet image and then applies its parameters.
 
We reconstructed $\mathbf{U}=\mathbf{F}^T\mathbf{P}$ and observed performance drop as it overfits. We get RMSE of $0.9305$ and $0.3762$ on enhancement and simulation data, respectively. We believe the real-world enhancement data has correlations naturally embedded in it unlike in synthetic data. Thus the performance drop is drastic in case of enhancement since the problem of recovering $\mathbf{P}$ only from $\mathbf{U}$ and $\mathbf{F}$ is ill-posed.

We also analyzed the effect of varying $\beta$ and $\delta$. Since our approach uses Bayesian probabilistic inference, small variations in $\beta$ and $\delta$ do not significantly affect the performance. Table \ref{tab:betaAndDelta} lists the various parameter settings and its effect on the performance of the second experiment (i.e. image enhancement): 

\begin{table}
\caption{Effect of varying $\beta$ and $\delta$} \label{tab:betaAndDelta} 
\vspace{4pt}
\begin{center}
    \begin{tabular}{ c c p{1cm}}
    \hline
    Parameter setting & RMSE (lower the better) \\ \hline \hline
    $\beta=0.001, \gamma=6$ & 0.3162 \\
    $\beta=0.01, \gamma=6$ & 0.0962 \\
    $\beta=0.02, \gamma=0.1$ & 0.0907 \\
    $\beta=0.2, \gamma=0.05$ & 0.0930 \\
    $\beta=0.8, \gamma=0.05$ & 0.0872 \\
    $\beta=0.1, \gamma=0.3$ & \textbf{0.0820} \\
    $\beta=0.1, \gamma=0.8$ & 0.0821 \\
    $\beta=0.1, \gamma=2$ & \textbf{0.0820} \\
    \hline \hline
    \end{tabular}
\end{center}
\end{table}

\section{Conclusion}

In this paper, we introduced a novel problem of predicting parameters of enhanced versions for a low-quality image by using its parameters and features. We developed an MF-inspired approach to solve this problem. We showed that by modeling the interactions across low-quality images, its parameters and its versions, we can outperform five state-of-art models in structured prediction and MF. We proposed inclusion of feature information into our formulation through a convex $\ell_{2,1}$-norm minimization, which works in an iterative fashion and is efficient. Thus our approach utilizes information which helps characterize input image. This leads to better generalization and prediction performance. Since other approaches do not model interdependence between image features and parameters of their corresponding enhanced versions, they start over-fitting quickly and produce an inferior prediction performance on the test set. Experiments on synthetic and real data demonstrated superiority of our approach over other state-of-art methods.

\textbf{Acknowledgement:} The work was supported in part by an ARO grant (\#W911NF1410371) and an ONR grant (\#N00014-15-1-2344). Any opinions expressed in this material are those of the authors and do not necessarily reflect the views of ARO or ONR.

{\small
\bibliographystyle{ieee}
\bibliography{egbib}
}

\newpage
\section{Supplementary}

The notation style is the same as that of the main paper. A different section is created to address each foot-note.

\section{Prior Distributions}

The prior distributions on $\mathbf{U,V}$ and $\mathbf{T}$ are chosen as normal distributions. We also consider a normal distribution to model the randomness in the attribute difference values $\Delta \mathbf{R}$. The details are as follows:

\begin{equation}
\begin{aligned}
&p(\Delta \mathbf{R}|\mathbf{U},\mathbf{V},\mathbf{T},\alpha)=\mathcal{N}_1(<U_i,V_j,T_k>,\alpha^{-1}) \\
&U_i \sim \mathcal{N}_D(0,\sigma_U^2 \mathbf{I_D}), \ \forall \ i=\{1,\hdots,N\} \\
&V_j \sim \mathcal{N}_D(0,\sigma_V^2 \mathbf{I_D}), \ \forall \ j=\{1,\hdots,M\} \\
&T_k \sim \mathcal{N}_D(0,\sigma_T^2 \mathbf{I_D}), \ \forall \ k=\{1,\hdots,K\},
\end{aligned}
\end{equation}

where $\alpha$ is precision, $\mathbf{I_D}$ is a $D \times D$ identity matrix, $\mathcal{N}_Z(\mathbf{\mu},\mathbf{\Lambda})$ is a $Z$-dimensional multivariate Gaussian distribution with $Z$-dimensional mean vector $\mu$ and a $Z \times Z$ covariance matrix $\mathbf{\Lambda}$. For both simulation and enhancement experiment, we use $\alpha=2$, $\sigma_U^2=\sigma_V^2=\sigma_T^2=0.01$.

We now choose prior distributions for the hyper-priors. 

\begin{equation}
\begin{aligned}
p(\alpha)= & \ \mathcal{W}(\alpha | \tilde{W}_0,\tilde{\nu}_0), \\
p(\Theta_U)= & \ p(\mu_U|\mathbf{\Lambda}_U) \cdot p(\mathbf{\Lambda}_U) \cdot \mathcal{N}(\mu_0,(\beta_0 \mathbf{\Lambda}_U)^{-1}) \cdot \\ \ & \mathcal{W}(\mathbf{\Lambda}_U|\mathbf{W}_0,\nu_0), \\
p(\Theta_V)= & \ p(\mu_V|\mathbf{\Lambda}_V) \cdot p(\mathbf{\Lambda}_V) \cdot \mathcal{N}(\mu_0,(\beta_0 \mathbf{\Lambda}_V)^{-1}) \cdot \\ \ & \mathcal{W}(\mathbf{\Lambda}_V|\mathbf{W}_0,\nu_0), \\
p(\Theta_T)= & \ p(\mu_T|\mathbf{\Lambda}_T) \cdot p(\mathbf{\Lambda}_T) \cdot \mathcal{N}(\mu_0,(\beta_0 \mathbf{\Lambda}_T)^{-1}) \cdot \\ \ & \mathcal{W}(\mathbf{\Lambda}_T|\mathbf{W}_0,\nu_0), \\
\end{aligned}
\end{equation}

\noindent Here, $\mathcal{W}$ is the Wishart distribution of a $D \times D$ random matrix $\mathbf{\Lambda}$ with $\nu_0$ degrees of freedom and a $D \times D$ scale matrix $\mathbf{W}_0$. The parameters in the hyper-priors: $\mu_0,\beta_0,\mathbf{W}_0,\nu_0,\tilde{W_0}$ and $\tilde{\nu_0}$ are treated as constants during training. They are set using prior knowledge of the application. For both experiments, we use: $\mu_0=0, \beta_0=1, \mathbf{W}_0=\mathbf{I}_D, \nu_0=D, \tilde{W}_0=1, \tilde{\nu}_0=1$. The Bayesian formulation of the factorization adjusts the parameters within a reasonable range.

\section{Sampling of Hyper-parameters}

\textbf{Conditional distributions in Gibbs Sampling:} The joint posterior distribution can be factorized as:

\begin{equation} \label{eq:Suppl_A1}
\begin{aligned}
p(\mathbf{U},\mathbf{V},\mathbf{T},\alpha,\Theta_U,\Theta_V,\Theta_T | \Delta \mathbf{R}) & \propto p(\Delta \mathbf{R} | \mathbf{U},\mathbf{V},\mathbf{T},\alpha) \cdot \\ \ & p(\mathbf{U} | \Theta_U) \cdot p(\mathbf{V} | \Theta_V) \cdot \\& p(\mathbf{T} | \Theta_T) \cdot p(\Theta_U) \cdot \\ \ & p(\Theta_V) \cdot p(\Theta_T) \cdot p(\alpha)
\end{aligned}
\end{equation}

We now derive the desired conditional distribution by substituting all the model components previously described.

\textbf{Hyper-parameters:} We use the conjugate prior for the parameter value precision $\alpha$, we have that the conditional distribution of $\alpha$ given $\Delta \mathbf{R,U,V}$ and $\mathbf{T}$ follows the Wishart distribution:

\begin{equation} \label{eq:Suppl_A2}
\begin{split}
& p(\alpha | \Delta \mathbf{R,U,V,T})=\mathcal{W}(\alpha | W_0^*,\nu_0^*), \\&
\nu_0^* = \tilde{\nu}_0+ \sum \limits_{i=1}^N \sum \limits_{j=1}^M \sum \limits_{k=1}^K I_{ij}^k, \\
(\tilde{W}_0^*)^{-1} & \ = \ \tilde{W}_0^{-1} + \\& \sum \limits_{i=1}^N \sum \limits_{j=1}^M \sum \limits_{k=1}^K (\Delta R_{ij}^k - <F_i^T \mathbf{P^*}+Q^*,V_j^*,T_k^*>)^2 \\
\end{split}
\end{equation}

\noindent where $I_{ij}^k=1$ if an attribute value $\Delta R_{ij}^k$ is present (not missing), otherwise $I_{ij}^k=0$. Also, $\mathbf{U}^*=F_i^T \mathbf{P^*}+Q^*$. For $\Theta_U=\{\mu_U,\mathbf{\Lambda_U\}}$, we can integrate out all the random variables given in Equation \ref{eq:Suppl_A1} except $\mathbf{U}$ and obtain the Gaussian-Wishart distribution:

\begin{equation} \label{eq:Suppl_A3}
\begin{aligned}
&p(\Theta_U | \mathbf{U})= \mathcal{N}(\mu_U | \mu_0^*,(\beta_0^* \mathbf{\Lambda}_U)^{-1}) \cdot \mathcal{W} (\mathbf{\Lambda}_U \vert \mathbf{W}_0^*,\nu_0^*),\\
&\mathbf{\mu}_0^*=\frac{\beta_0 \mathbf{\mu}_0 + N \bar{U}}{\beta_0+N}, \beta_0^* = \beta_0+N, \nu_0^*=\nu_0+N; \\
&(\mathbf{W}_0^*)^{-1}=\mathbf{W}_0^{-1} + N \bar{\mathbf{S}} + \frac{\beta_0 N}{\beta_0+N} \cdot (\mu_0-\bar{U})(\mu_0-\bar{U})^T,\\
& \text{where,} \ \ \bar{U}=\frac{1}{N} \sum \limits_{i=1}^N U_i, \ \bar{\mathbf{S}}=\frac{1}{N} \sum \limits_{i=1}^N (U_i-\bar{U})(U_i-\bar{U})^T.
\end{aligned}
\end{equation}

similarly, $\Theta_V=\{\mu_V,\mathbf{\Lambda}_V\}$ is conditionally independent of all other parameters given $\mathbf{V}$, and its conditional distribution has the form:

\begin{equation} \label{eq:Suppl_A4}
\begin{split}
&p(\Theta_V | \mathbf{V})= \mathcal{N}(\mu_V | \mu_0^*,(\beta_0^* \mathbf{\Lambda}_V)^{-1}) \cdot \mathcal{W} (\mathbf{\Lambda}_V \vert \mathbf{W}_0^*,\nu_0^*),\\
&\mathbf{\mu}_0^*=\frac{\beta_0 \mathbf{\mu}_0 + N \bar{V}}{\beta_0+N}, \beta_0^* = \beta_0+N, \nu_0^*=\nu_0+N; \\
&(\mathbf{W}_0^*)^{-1}=\mathbf{W}_0^{-1} + N \bar{\mathbf{S}} + \frac{\beta_0 N}{\beta_0+N} \cdot (\mu_0-\bar{V})(\mu_0-\bar{V})^T,\\
&\bar{V}=\frac{1}{N} \sum \limits_{i=1}^N V_i, \ \bar{\mathbf{S}}=\frac{1}{N} \sum \limits_{i=1}^N (V_i-\bar{V})(V_i-\bar{V})^T.
\end{split}
\end{equation}

similarly, $\Theta_T=\{\mu_T,\mathbf{\Lambda}_T\}$ is conditionally independent of all other parameters given $\mathbf{T}$, and its conditional distribution has the form:

\begin{equation} \label{eq:Suppl_A5}
\begin{split}
&p(\Theta_T | \mathbf{T})= \mathcal{N}(\mu_T | \mu_0^*,(\beta_0^* \mathbf{\Lambda}_T)^{-1}) \cdot \mathcal{W} (\mathbf{\Lambda}_T \vert \mathbf{W}_0^*,\nu_0^*),\\
&\mathbf{\mu}_0^*=\frac{\beta_0 \mathbf{\mu}_0 + N \bar{T}}{\beta_0+N}, \beta_0^* = \beta_0+N, \nu_0^*=\nu_0+N; \\
&(\mathbf{W}_0^*)^{-1}=\mathbf{W}_0^{-1} + N \bar{\mathbf{S}} + \frac{\beta_0 N}{\beta_0+N} \cdot (\mu_0-\bar{T})(\mu_0-\bar{T})^T,\\
&\bar{T}=\frac{1}{N} \sum \limits_{i=1}^N T_i, \ \bar{\mathbf{S}}=\frac{1}{N} \sum \limits_{i=1}^N (T_i-\bar{T})(T_i-\bar{T})^T.
\end{split}
\end{equation}

\textbf{Model Parameters:} We first consider the latent example (data sample) features $\mathbf{U}$. Since its columns affect the example features independently, its conditional distribution factorizes w.r.t. individual $U_i$.

\begin{equation}
p(\mathbf{U} | \Delta \mathbf{R,V,T},\alpha,\Theta) = \prod \limits_{i=1}^N p(U_i | \Delta \mathbf{R,V,T},\alpha,\Theta_U).
\end{equation}

Then for each latent example feature vector $\mathbf{U}_i$,

\begin{equation}
\begin{split}
&p(U_i | \Delta \mathbf{R,V,T},\alpha,\Theta_U)=\mathcal{N}(U_i | \mu_i^*,(\mathbf{\Lambda}_i^*)^{-1}), \\&
\mu_i^* \equiv (\mathbf{\Lambda}_i^*)^{-1} (\mathbf{\Lambda}_U \mu_U + \alpha \sum \limits_{j=1}^M \sum \limits_{k=1}^K I_{ij}^k R_{ij}^k Y_{jk}) \\&
\mathbf{\Lambda}_i^{*} \equiv \mathbf{\Lambda}_U+\alpha \sum \limits_{k=1}^K \sum \limits_{j=1}^M I_{ij}^k Y_{jk} Y_{jk}^T.
\end{split}
\end{equation}

\noindent where $Y_{jk} \equiv V_j \cdot T_k$, which represents element-wise product between $V_j$ and $T_k$.     

Similarly, for each latent modified version feature $V_j$, we have:

\begin{equation}
\begin{split}
&p(V_j | \Delta \mathbf{R,U,T},\alpha,\Theta_V)=\mathcal{N}(V_j | \mu_j^*,(\mathbf{\Lambda}_j^*)^{-1}), \\&
\mu_j^* \equiv (\mathbf{\Lambda}_j^*)^{-1} (\mathbf{\Lambda}_V \mu_V + \alpha \sum \limits_{i=1}^N \sum \limits_{k=1}^K I_{ij}^k R_{ij}^k Y_{ik}) \\&
\mathbf{\Lambda}_j^{*} \equiv \mathbf{\Lambda}_V+\alpha \sum \limits_{k=1}^K \sum \limits_{j=1}^M I_{ij}^k Y_{ik} Y_{ik}^T.
\end{split}
\end{equation}

\noindent where $Y_{ik} \equiv (F^T_i \mathbf{P}+Q) \cdot T_k$

For each latent attribute feature $T_k$, we have:

\begin{equation}
\begin{split}
&p(T_k | \Delta \mathbf{R,U,V},\alpha,\Theta_T)=\mathcal{N}(T_k | \mu_k^*,(\mathbf{\Lambda}_k^*)^{-1}), \\&
\mu_k^* \equiv (\mathbf{\Lambda}_k^*)^{-1} (\mathbf{\Lambda}_T \mu_T + \alpha \sum \limits_{i=1}^N \sum \limits_{j=1}^M I_{ij}^k R_{ij}^k Y_{ij}) \\&
\mathbf{\Lambda}_k^{*} \equiv \Lambda_T+\alpha \sum \limits_{k=1}^K \sum \limits_{j=1}^M I_{ij}^k Y_{ij} Y_{ij}^T.
\end{split}
\end{equation}

\noindent where $Y_{ij} \equiv (F^T_i \mathbf{P}+Q) \cdot V_j$


\end{document}